\documentclass[letterpaper,journal]{IEEEtran}

\usepackage{amsmath,amssymb,amsfonts}
\usepackage{pifont}
\usepackage{array}
\usepackage{booktabs}
\usepackage[caption=false,font=normalsize,labelfont=sf,textfont=sf]{subfig}
\usepackage{textcomp}
\usepackage{stfloats}
\usepackage{url}
\usepackage{graphicx}
\usepackage{cite}

\graphicspath{{./}{./figs/}{./figs/rainds_selected_semantic_v3/best_psnr/}}

\newcolumntype{L}{l}
\newcolumntype{C}{c}
\newcommand{\tblwidth}{\linewidth}
\newcommand{\cmark}{\checkmark}
\newcommand{\xmark}{\times}

\newcommand{\safeincludegraphics}[2][]{%
  \IfFileExists{#2}{\includegraphics[#1]{#2}}{%
    \fbox{\parbox[c][4cm][c]{0.9\linewidth}%
      {\centering\small Missing: \texttt{\detokenize{#2}}}}%
  }%
}

\begin{document}
\title{AFRNet: Adaptive Frequency Redistribution Network for Image Deraining}

\author{Dezhi Zhang, Shangrong Yang, Husheng Guo, Zhijie Shen, and Wenjian Wang%
\thanks{This work was supported in part by the National Natural Science
Foundation of China under Grants U21A20513 and 62476157. Corresponding
authors: Shangrong Yang and Wenjian Wang.}%
\thanks{Dezhi Zhang, Shangrong Yang, Husheng Guo, and Wenjian Wang are with the School
of Computer and Information Technology, Shanxi University, Taiyuan 030006,
China (e-mail: zhangdezhi1@sxu.edu.cn; sryang@sxu.edu.cn;
guohusheng@sxu.edu.cn;
wjwang@sxu.edu.cn).}%
\thanks{Zhijie Shen is with Beijing Jiaotong University, Beijing, China
(e-mail: zjshen@bjtu.edu.cn).}}

\markboth{IEEE Transactions on Multimedia}%
{Zhang \MakeLowercase{\textit{et al.}}: AFRNet: Adaptive Frequency Redistribution Network for Image Deraining}

\maketitle

\begin{abstract}
Image deraining aims to recover rain-free images from rain-degraded observations. Recent frequency-aware deraining methods have demonstrated promising performance by removing rain artifacts in the spectral domain. However, since rain artifacts and background textures often appear in several correlated subbands, prior frequency-domain deraining models still enhance frequency subbands separately. Such band-wise processing limits information exchange across subbands. Without context from other subbands, rain and clean content that appear together in one subband cannot be separated. This leads to either residual rain streaks or over-smoothing of fine details. To address this limitation, we propose AFRNet, an Adaptive Frequency Redistribution Network for image deraining. First, we introduce an Adaptive Frequency Recombination Module (AFRM) that performs cross-band modulation among wavelet subbands, enabling information interaction and adaptive recombination of degraded frequency components. We then develop a Subband Compensation Block (SCB) that compensates for decoder subband degradation by leveraging encoder subband features through inter-subband attention in the wavelet domain, alleviating detail loss during reconstruction. Extensive experiments demonstrate that AFRNet achieves competitive performance and consistent results across diverse deraining benchmarks. 
\end{abstract}

\begin{IEEEkeywords}
Image deraining, adaptive frequency recombination, wavelet subband modulation, subband information compensation, compound rain removal.
\end{IEEEkeywords}

\section{Introduction}

Rain substantially degrades image quality and reduces the reliability of outdoor vision systems for tasks such as object detection, semantic segmentation, and depth estimation~\cite{hnewa2021object,pfeuffer2019robust,wang2024weatherdepth}.
Early methods relied on hand-crafted priors and signal models, such as image decomposition, sparse coding, and layer-based restoration~\cite{kang2011automatic,luo2015removing,eigen2013restoring}, while CNN-based models~\cite{yang2017deep,ren2019progressive,fu2017clearing,li2018recurrent} improved restoration by learning hierarchical mappings from paired data. Transformer-based architectures~\cite{zamir2022restormer,liang2021swinir,wang2022uformer,tu2022maxim} further addressed the limited receptive field of CNNs by modeling long-range spatial dependencies. More recently, frequency-aware deraining methods~\cite{hsu2023recurrent,wen2024heavy,ali2024wavelet} have emerged because rain streaks exhibit distinct spectral characteristics that are difficult to distinguish in the spatial domain, enabling more effective separation.

\begin{figure}[!t]
  \centering
  \safeincludegraphics[width=\columnwidth]{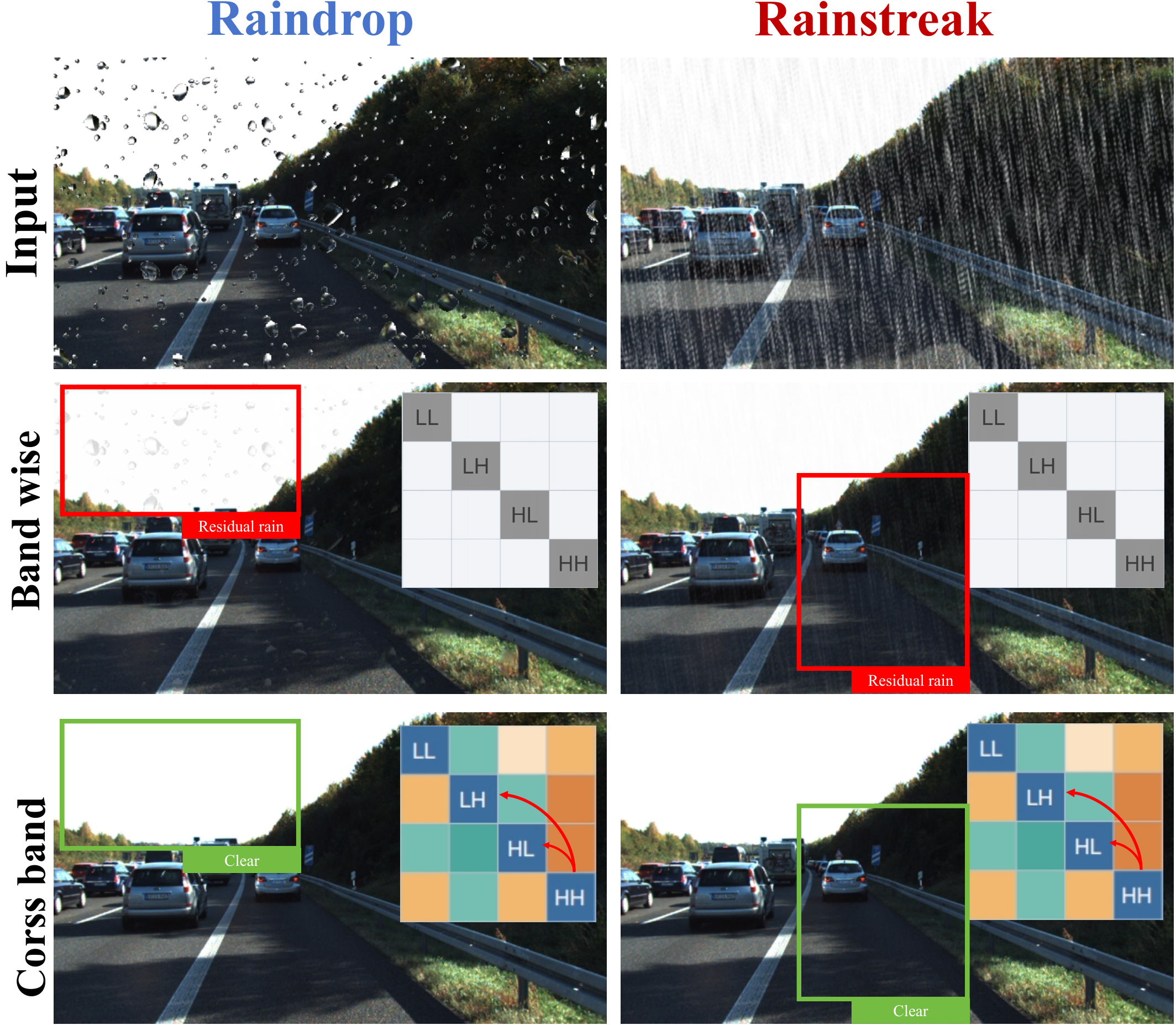}
  \caption{Band-wise processing vs.\ cross-band redistribution.
    Band-wise processing retains residual artifacts, while cross-band
    redistribution enables cleaner restoration. Insets show subband
    interaction patterns (diagonal: isolated processing; off-diagonal:
    cross-band exchange).}
  \label{fig:bandwise_vs_crossband}
\end{figure}

Despite these advances, many existing frequency-domain deraining models still enhance individual frequency bands with limited interaction among them. This limitation becomes critical when rain artifacts and background textures are distributed over correlated spectral components. In this case, corrupted bands are reconstructed only from their own degraded representations, without leveraging information from other subbands. As a result, rain artifacts may remain within degraded components, while clean structures may suffer unnecessary distortion during reconstruction. This limitation is illustrated in Fig.~\ref{fig:bandwise_vs_crossband}.

\begin{figure*}[!t]
  \centering
  \safeincludegraphics[width=\textwidth]{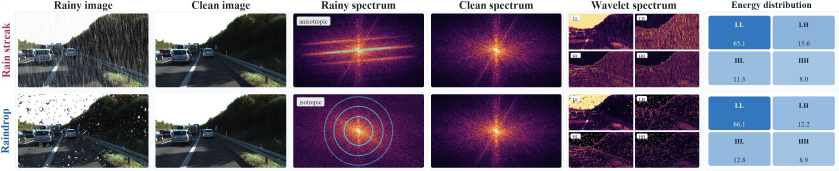}
  \caption{Frequency analysis of rain degradation. Rain streaks and raindrops
    show different frequency properties: rain streaks are more anisotropic,
    whereas raindrops are more isotropic. Single-band processing cannot handle
    both types well, which motivates cross-band modulation.}
  \label{fig:frequency_analysis}
\end{figure*}

Fig.~\ref{fig:frequency_analysis} provides frequency-domain evidence for
the above limitation. Different rain forms show distinct spectral behaviors:
rain streaks are more direction-dependent, whereas raindrops exhibit weaker
directional spectral responses. In the wavelet domain, rain artifacts and
background textures are not confined to a single subband, but spread across
correlated subband components. The subband energy statistics further show that
rain-layer responses are not restricted to high-frequency components. These
observations suggest that fixed band-wise enhancement is insufficient for
complex rain degradation, and that frequency modeling should consider both
cross-band relations and low-frequency responses.

Viewed from this perspective, many frequency-domain methods remain
constrained by three critical limitations: (1) they assume rain follows a single frequency distribution pattern, thereby enhancing individual subbands independently; (2) they process correlated subbands separately, limiting the use of complementary cross-subband information; and (3) their pipelines lack the ability to leverage encoder subband information. Details lost during downsampling and upsampling are therefore not compensated. This limits the recovery of local structures and fine details.

We propose the Adaptive Frequency Redistribution Network (AFRNet) for image deraining. The Adaptive Frequency Recombination Module (AFRM) performs adaptive cross-band modulation and recombination on wavelet subbands. The Subband Compensation Block (SCB) compensates for degradation in decoder subbands by leveraging encoder subband information through inter-subband attention. AFRNet achieves a balance between rain artifact removal and texture preservation across diverse rain scenarios. The contributions are summarized as follows:
\begin{itemize}
\item We introduce an Adaptive Frequency Recombination Module (AFRM) that performs cross-band modulation across wavelet subbands and recombines their features, enabling degraded frequency components to leverage complementary information from correlated subbands.
\item We introduce inter-subband attention to leverage encoder subband
  features for compensating decoder subbands, forming a Subband Compensation
  Block (SCB) that restores lost structural and high-frequency details.
\item Extensive experiments on synthetic, paired real-rain, and compound rain
  benchmarks verify the effectiveness and generalization of the
  proposed method.
\end{itemize}

\section{Related Work}

\subsection{Spatial-Domain Deraining Methods}

\noindent\textbf{CNN-based deraining methods.} These data-driven methods learn rain removal directly in the image domain. JORDER~\cite{yang2017deep} jointly detects rain streaks and restores the clean background, showing that explicit rain analysis can improve single-image deraining. Later CNN restorers further improve spatial refinement through recurrent prediction, context aggregation, and multi-stage restoration~\cite{fu2017clearing,zhang2018density,ren2019progressive,li2018recurrent,zamir2021multi}. MPRNet~\cite{zamir2021multi} is a representative multi-stage model that progressively restores degraded images and preserves spatial details. However, these methods mainly separate rain and background according to
spatial appearance. When rain artifacts have edge-like patterns similar to clean textures, reliable separation remains difficult.

\noindent\textbf{Attention-based spatial methods.} Transformer-based and
unified restoration methods enlarge the effective context through attention,
sparse interaction, and prompt conditioning~\cite{zamir2022restormer,liang2021swinir,wang2022uformer,tu2022maxim,mou2022colanet,Chen_2023_CVPR_DRSformer,chen2023udrs2former,chen2024bidirectional,li2022all,Potlapalli2023PromptIR,valanarasu2022transweather,zhu2023wgws}.
Restormer~\cite{zamir2022restormer} and DRSformer~\cite{Chen_2023_CVPR_DRSformer} are representative models in this line,
showing the effectiveness of attention-based spatial modeling for image
restoration and deraining. These methods greatly improve spatial
restoration quality because attention enlarges the region from which
features can be aggregated. Nevertheless, their restoration is still mainly
driven by spatial attention, prompts, or refinement operations. When rain
artifacts and clean textures have similar spatial patterns, this
spatial-only restoration has limited information for judging which
structures should be changed. As a result, restoration strength tends to
be applied in a similar way across the image, which can leave artifacts
in heavily degraded regions or over-process clean structures.

\subsection{Frequency-Aware Restoration Methods}

\noindent\textbf{Frequency-based methods.} These methods model rain degradation through spectral representations. Wavelet-based methods explicitly decompose image features into subbands and preserve structural information through frequency-aware restoration~\cite{hsu2023recurrent,wen2024heavy}. FADformer~\cite{gao2024efficient} uses frequency-domain learning and contrastive regularization for efficient rain removal, while AFENet~\cite{yan2024afenet} enhances deraining through adaptive frequency processing. These methods show that spectral representations provide an effective perspective for separating directional and scale-related rain patterns. However, most frequency operations are still used for enhancement, selection, fusion, or regularization within predefined processing streams. They rarely allow the input to guide information exchange among subbands before reconstruction. Thus, frequency redistribution remains insufficiently studied for complex rain degradation, especially when useful information is distributed across related subbands.

Recent restoration and deraining methods further combine frequency modeling with adaptive restoration, degradation-content decoupling, frequency-prioritized attention, and state-space modeling~\cite{cui2025adair,wang2025ddcnet,dong2026high,yamashita2024dfssm,li2025msrainmamba,zhu2025derainmamba,xue2025prism,zhang2026multiscale,liu2025funet,huang2024wavedm,wang2026niinet}.
For example, AdaIR~\cite{cui2025adair} mines frequency information for
adaptive all-in-one restoration, while FreqMamba~\cite{zou2024freqmamba},
FourierMamba~\cite{li2024fouriermamba}, MS-RainMamba~\cite{li2025msrainmamba}, and DeRainMamba~\cite{zhu2025derainmamba} extend
state-space modeling toward frequency-aware deraining. These methods improve feature extraction and long-range degradation modeling. However, frequency information is mainly used to enrich feature representations or model long-range dependencies, rather than to explicitly exploit the relationships among wavelet subbands. In contrast, AFRNet employs the Adaptive Frequency Recombination Module (AFRM) to adaptively modulate and recombine features across wavelet subbands, while the Subband Compensation Block (SCB) compensates for decoder subband degradation by leveraging encoder subband features through inter-subband attention in the wavelet domain.

\section{Method}
\label{sec:method}
\subsection{Overall Framework}
\label{sec:overall}

\begin{figure*}[!t]
  \centering
  \safeincludegraphics[width=\textwidth]{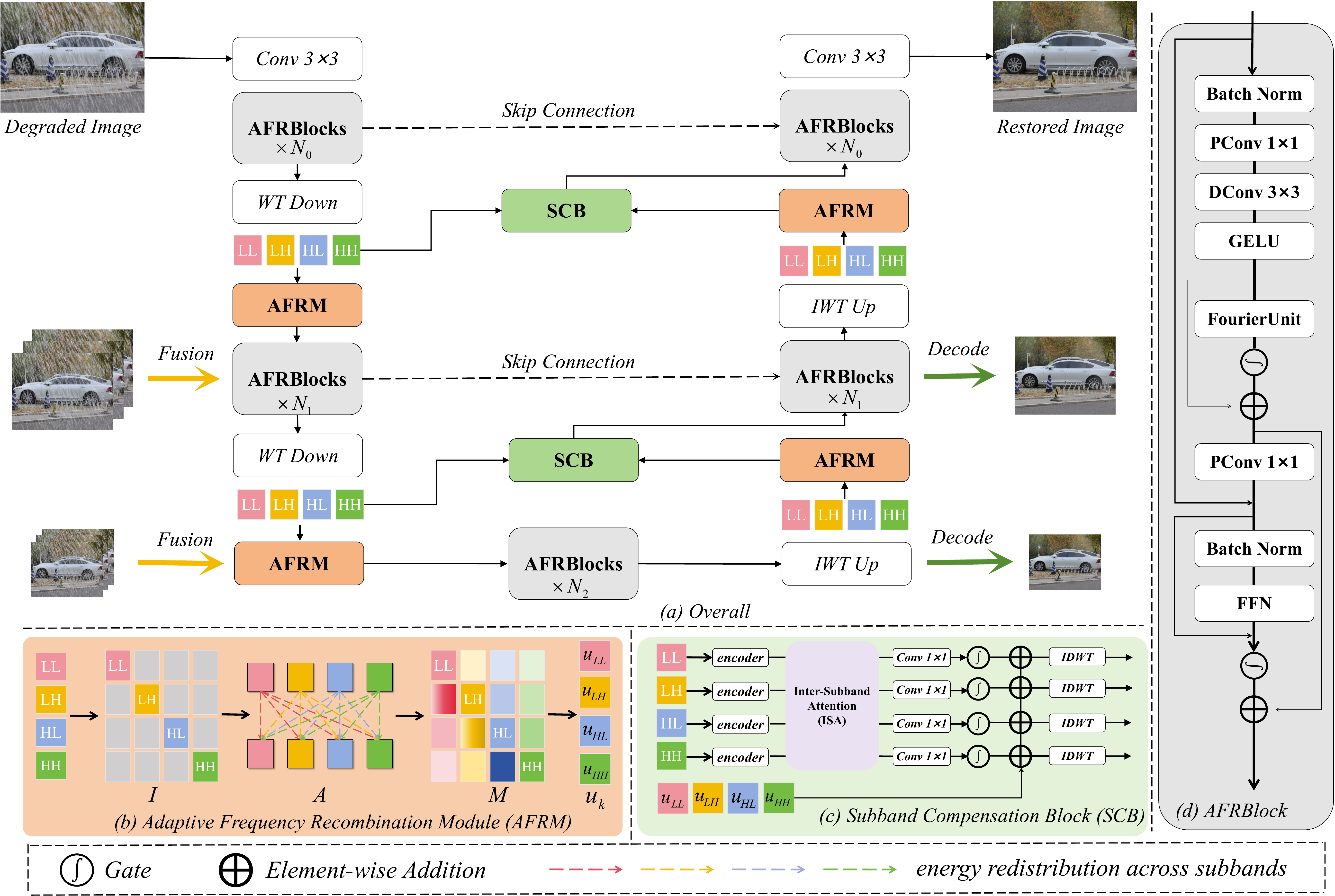}
  \caption{Architecture of AFRNet for image deraining. AFRNet combines
    frequency redistribution with spatial feature restoration. Its main
    components are: (a) the Adaptive Frequency Recombination Module (AFRM),
    which performs content-adaptive cross-subband modulation to redistribute
    frequency information; (b) the Subband Compensation Block (SCB), which
    transfers encoder subband features to decoder subbands to recover details
    lost during reconstruction; and (c) the Adaptive Frequency Restoration
    Block (AFRBlock), which combines local spatial processing with standard
    Fourier mixing to refine spatial features. The wavelet encoder-decoder
    provides the feature hierarchy and subband interfaces for these
    operations. WT Down and IWT Up use the wavelet transform and inverse
    wavelet transform to replace convolution-based downsampling and
    upsampling.}
  \label{fig:architecture}
\end{figure*}

AFRNet consists of a shallow feature extractor, a hierarchical encoder-decoder
backbone, the Adaptive Frequency Recombination Module (AFRM), the Subband
Compensation Block (SCB), and the Adaptive Frequency Restoration Block
(AFRBlock). AFRM is the frequency-side core and performs content-adaptive
redistribution across correlated wavelet subbands, allowing rain-corrupted
frequency components to exploit complementary structural information. AFRBlock
is the spatial-side core and restores spatial features through local spatial
processing and standard Fourier mixing. SCB is a supporting decoder module
that uses encoder subband information to recover details weakened during
reconstruction. We first describe the overall data flow and then present
AFRM, SCB, AFRBlock, and the loss function.

As shown in Fig.~\ref{fig:architecture}, a rainy image $I_r$ is first mapped to
the initial feature $X_0$ by the shallow feature extractor. The encoder then
processes the feature through the hierarchical backbone. At each encoder
downsampling site, the wavelet transform decomposes the feature into the four
subbands $LL$, $LH$, $HL$, and $HH$, after which AFRM redistributes their
information before the feature is passed to the next level. AFRBlock operates
on the spatial features at each level to refine local structures and model
global feature interactions. During decoding, the decoder feature is mapped to
four subbands and processed by AFRM. SCB uses the corresponding encoder
subband features to compensate the decoder subbands, and the inverse wavelet
transform reconstructs the spatial feature. Finally, AFRNet predicts a
residual image and adds it to $I_r$ to obtain the clean image.

\subsection{Adaptive Frequency Recombination Module (AFRM)}
\label{sec:afrm}

To address the limited interaction among wavelet subbands, we introduce AFRM
to adaptively combine their features before reconstruction. Let
$\mathcal{B}=\{LL,LH,HL,HH\}$ denote the four wavelet subbands, and let
$X^k\in\mathbb{R}^{C\times H\times W}$ be the feature of subband $k\in
\mathcal{B}$. AFRM concatenates the four subband features, predicts mixing
weights on a coarse spatial grid with a $1\times1$ convolution, normalizes the
weights over the source-subband dimension with softmax, and upsamples them to
the subband resolution. The resulting $\mathbf{A}$ is a content-dependent
mixing matrix whose entries specify how each source subband contributes to an
output subband.

To retain the original subband representation at initialization, AFRM combines
the learned mixing with the $4\times4$ identity matrix:
\begin{equation}
  \mathbf{M}=(1-\alpha)\mathbf{I}+\alpha\mathbf{A},
  \label{eq:afrm_matrix}
\end{equation}
where $\mathbf{M}$ is the final adaptive recombination matrix, $\mathbf{I}$ is
the identity matrix, $\mathbf{A}$ is the content-dependent mixing matrix, and
$\alpha$ is a learnable scalar that controls the strength of adaptive
recombination. We initialize $\alpha$ to zero and constrain it to $[0,1]$
during training. The output feature of subband $i\in\mathcal{B}$ is computed
as
\begin{equation}
  U^i=\sum_{j\in\mathcal{B}}M_{ij}X^j,
  \label{eq:afrm_recombination}
\end{equation}
where $M_{ij}$ is the contribution of source subband $j$ to output subband $i$.
Therefore, each output subband can use complementary information from the other
subbands instead of relying only on its own degraded feature. AFRM modifies the
current wavelet representation in the main path and does not use encoder
features as an external compensation signal.

\subsection{Subband Compensation Block (SCB)}
\label{sec:scb}

Although AFRM enables adaptive interaction among the current wavelet subbands,
it does not recover details that may be lost during encoder-decoder resolution
changes. We therefore introduce SCB to transfer encoder subband information to
the decoder. SCB is inserted after decoder-side AFRM and before inverse wavelet
reconstruction. For each $k\in\mathcal{B}$, the encoder feature $S_l^k$ at
scale $l$ is mapped to a hidden feature $E_l^k$ by a band-specific
$1\times1$ convolution and GELU activation. The four hidden features retain
their subband identities and are then processed jointly by inter-subband
attention.

The inter-subband attention (ISA) treats the four wavelet subbands as tokens,
rather than treating spatial locations as tokens. Let
$E_l=\operatorname{Concat}_{k\in\mathcal{B}}(E_l^k)$. Query, key, and value
features are generated from $E_l$ by a pointwise convolution followed by a
depthwise convolution:
\begin{equation}
  [Q_l,K_l,V_l]=
  \operatorname{DWConv}_{3\times3}
  \bigl(\operatorname{Conv}_{1\times1}(E_l)\bigr),
  \label{eq:isa_qkv}
\end{equation}
where $Q_l$, $K_l$, and $V_l$ denote the query, key, and value features,
respectively. After channel splitting into attention heads, the features of
each subband are flattened over their channel and spatial dimensions. Thus,
ISA computes a $4\times4$ attention matrix for each head and image. With
$\operatorname{Norm}(\cdot)$ denoting $\ell_2$ normalization along the
flattened feature dimension, the attention matrix is
\begin{equation}
  A_l^{\mathrm{ISA}}=\operatorname{Softmax}\!\left(
  \tau_l\,\operatorname{Norm}(Q_l)\operatorname{Norm}(K_l)^{\mathsf T}
  \right),
  \label{eq:isa_attention}
\end{equation}
where $\tau_l$ is a learnable temperature parameter. The attention weights are
normalized over the source-subband dimension and are applied to $V_l$. ISA
therefore exchanges information among the four subbands while using their
spatial content to determine the strength of each cross-subband interaction;
it does not construct an $HW\times HW$ spatial attention matrix.

The attended features are split into four subbands and projected to the decoder
channel dimension. Let $U_l^k$ denote the decoder feature of subband $k$ after
AFRM, and let $R_l^k$ denote the corresponding projected ISA output. SCB
injects the encoder-derived correction with a learnable band-specific strength:
\begin{equation}
  \widetilde U_l^k=U_l^k+\beta_l^kR_l^k,
  \qquad k\in\{LL,LH,HL,HH\}.
  \label{eq:scb_injection}
\end{equation}
Here $\beta_l^k$ controls the compensation strength for subband $k$. The
coefficients $\beta_l^k$ and the ISA output projection are initialized to zero,
so SCB starts as an identity path and learns which subbands require encoder
guidance. The compensated subbands are then passed to inverse wavelet
reconstruction. In contrast to AFRM, SCB does not recombine the decoder
representation directly; it adds a residual detail correction derived from
encoder subband features.

\subsection{Adaptive Frequency Restoration Block (AFRBlock)}
\label{sec:afrblock}

AFRBlock is the basic feature restoration unit in the encoder and decoder. It
operates on spatial features and is independent of the wavelet-domain exchange
performed by AFRM and SCB. Given an input feature
$X\in\mathbb{R}^{C\times H\times W}$, the token-mixing path first applies a
pointwise convolution and a depthwise convolution to introduce local spatial
context. It then expands the channels with a pointwise convolution, applies the
FourierUnit, and projects the result back to $C$ channels. The FourierUnit
performs the standard sequence of a two-dimensional real FFT, pointwise
processing of the real and imaginary parts, and an inverse real FFT. This
provides global feature mixing with low computational cost.

The Fourier-mixing output is followed by a local feed-forward transformation,
which expands the channels, applies a depthwise $3\times3$ convolution and
GELU activation, and projects the features back to $C$ channels. Let
$\mathcal{M}_F$ denote the complete token-mixing path and let
$\operatorname{Norm}(\cdot)$ denote the feature normalization used in the
block. AFRBlock is written as
\begin{equation}
  \begin{aligned}
  X'&=X+\beta\mathcal{M}_F(\operatorname{Norm}(X)),\\
  X_{\mathrm{out}}&=X'+\gamma\operatorname{FFN}(\operatorname{Norm}(X')),
  \end{aligned}
  \label{eq:afrblock}
\end{equation}
where $X'$ is the feature after token mixing, $X_{\mathrm{out}}$ is the block
output, and $\beta$ and $\gamma$ are learnable channel-wise residual scales.
The FourierUnit supplies global mixing, while the depthwise convolutions and
FFN refine local structures. AFRBlock therefore forms the spatial restoration
backbone, whereas AFRM and SCB operate on the wavelet representation.

\subsection{Loss Function}
\label{sec:loss}

AFRNet produces two image outputs and two outputs aligned with the first-level
wavelet representation. Let $\mathcal{P}$ denote the set of four
prediction-target pairs, and let $Z$ and $\overline Z$ denote a prediction and
its corresponding target. We use an image-domain $L_1$ loss together with an
FFT-domain loss. Let $\mathcal{C}(Z)$ be the concatenation of the real and
imaginary parts of the two-dimensional FFT of $Z$. The loss is defined as
\begin{equation}
  \begin{aligned}
  \mathcal{L}_1(Z,\overline Z)&=\|Z-\overline Z\|_1,\\
  \mathcal{L}_{F}(Z,\overline Z)&=
  \|\mathcal{C}(Z)-\mathcal{C}(\overline Z)\|_1,\\
  \mathcal{L}&=\sum_{(Z,\overline Z)\in\mathcal{P}}
  \left[\mathcal{L}_1(Z,\overline Z)+0.1\mathcal{L}_{F}(Z,\overline Z)\right].
  \end{aligned}
  \label{eq:loss}
\end{equation}
The $L_1$ term constrains pixel-level fidelity, while the Fourier term
preserves the frequency structure of the clean target in the image and
intermediate wavelet outputs.
\section{Experiments}

\subsection{Experimental Settings}
\label{sec:results}

We evaluate AFRNet under three protocols. Under the Rain13K protocol,
Rain13K~\cite{yang2017deep} is used for training, and the model is tested on
Test100, Rain100H, Rain100L, Test2800, and Test1200~\cite{zhang2018density}.
We independently train and evaluate models on Rain200L/H,
SPA-Data~\cite{Wang_2019_CVPR}, and RainDS. For RainDS, we present
quantitative results on the synthetic RS, RD, and RDRS subsets and provide
qualitative results on the unpaired RainDS-Real set. PSNR and SSIM are
computed on the luminance (Y) channel for paired test sets.

For training, Rain13K uses $64\times64$ random crops, whereas the other
datasets use $256\times256$ random crops. We apply random horizontal flips
and $90^{\circ}$ rotations for augmentation. All AFRNet variants are trained
for 300K iterations with AdamW, a batch size of 8, and a weight decay of
$10^{-4}$. The learning rate is initialized to $8\times10^{-4}$ and reduced
to $10^{-6}$ using cosine annealing with a short warm-up. AFRNet-S and
AFRNet-L use $(N_0,N_1,N_2)=(6,4,2)$ and $(8,6,4)$, respectively. All variants
use the loss described in Sec.~\ref{sec:loss}.

\subsection{Quantitative Results}

\begin{table*}[!t]
\caption{Quantitative Comparison on the Five Rain13K Test Sets in Terms of
  Y-Channel PSNR and SSIM. Best and second-best results are shown in
  \textbf{bold} and \underline{underlined}, respectively.}
\label{tab:rain13k}
\centering
\scriptsize
\setlength{\tabcolsep}{1.4pt}
\begin{tabular*}{\tblwidth}{@{\extracolsep{\fill}}LC*{12}{C}@{}}
\toprule
Method & Year
  & \multicolumn{2}{c}{Test100}
  & \multicolumn{2}{c}{Rain100H}
  & \multicolumn{2}{c}{Rain100L}
  & \multicolumn{2}{c}{Test2800}
  & \multicolumn{2}{c}{Test1200}
  & \multicolumn{2}{c}{Average} \\
\cmidrule(lr){3-4}
\cmidrule(lr){5-6}
\cmidrule(lr){7-8}
\cmidrule(lr){9-10}
\cmidrule(lr){11-12}
\cmidrule(lr){13-14}
  & & PSNR $\uparrow$ & SSIM $\uparrow$
  & PSNR $\uparrow$ & SSIM $\uparrow$
  & PSNR $\uparrow$ & SSIM $\uparrow$
  & PSNR $\uparrow$ & SSIM $\uparrow$
  & PSNR $\uparrow$ & SSIM $\uparrow$
  & PSNR $\uparrow$ & SSIM $\uparrow$ \\
\midrule
RESCAN~\cite{li2018recurrent}       & ECCV2018  & 21.59 & 0.726 & 18.01 & 0.467 & 24.15 & 0.791 & 24.50 & 0.765 & 24.40 & 0.759 & 22.53 & 0.702 \\
PReNet~\cite{ren2019progressive}    & CVPR2019  & 23.17 & 0.752 & 17.63 & 0.487 & 27.76 & 0.876 & 27.20 & 0.825 & 26.05 & 0.792 & 24.36 & 0.746 \\
SPDNet~\cite{yi2021spdnet}          & ICCV2021  & 24.25 & 0.848 & 25.87 & 0.809 & 28.63 & 0.880 & 31.05 & 0.904 & 30.42 & 0.893 & 28.04 & 0.867 \\
PCNet~\cite{jiang2021pcnet}         & TIP2021   & 23.29 & 0.762 & 20.83 & 0.563 & 26.64 & 0.817 & 27.10 & 0.818 & 26.53 & 0.791 & 24.88 & 0.750 \\
MPRNet~\cite{zamir2021multi}        & CVPR2021  & 25.66 & 0.859 & 28.23 & 0.850 & 31.94 & 0.930 & 32.14 & 0.925 & 31.32 & 0.901 & 29.86 & 0.893 \\
HINet~\cite{chen2021hinet}          & CVPRW2021 & 23.21 & 0.767 & 20.85 & 0.598 & 27.03 & 0.842 & 28.56 & 0.843 & 27.77 & 0.821 & 25.48 & 0.774 \\
DANet~\cite{jiang2022danet}         & IJCAI2022 & 23.96 & 0.839 & 23.00 & 0.791 & 29.51 & 0.906 & 30.32 & 0.903 & 29.99 & 0.888 & 27.36 & 0.865 \\
Uformer~\cite{wang2022uformer}       & CVPR2022  & 23.87 & 0.815 & 22.43 & 0.700 & 28.39 & 0.883 & 29.71 & 0.886 & 28.65 & 0.856 & 26.61 & 0.828 \\
ALformer~\cite{jiang2022magicelf}    & ACMMM2022 & 24.41 & 0.844 & 25.10 & 0.807 & 29.39 & 0.903 & 31.36 & 0.916 & 30.40 & 0.897 & 28.13 & 0.874 \\
NAFNet~\cite{chen2022nafnet}         & ECCV2022  & 25.75 & 0.845 & 26.76 & 0.813 & 31.27 & 0.925 & 31.71 & 0.918 & 30.62 & 0.892 & 29.22 & 0.879 \\
MFDNet~\cite{wang2023mfdnet}         & TIP2024   & 25.90 & 0.870 & 27.06 & 0.850 & 32.76 & 0.944 & 31.92 & 0.925 & 31.15 & 0.909 & 29.76 & 0.899 \\
HCT-FFN~\cite{chen2023hctffn}        & AAAI2023  & 24.86 & 0.847 & 26.70 & 0.819 & 29.94 & 0.906 & 31.46 & 0.915 & 31.23 & 0.901 & 28.84 & 0.878 \\
DRSformer~\cite{Chen_2023_CVPR_DRSformer} & CVPR2023  & 27.86 & 0.885 & 28.16 & 0.864 & 34.79 & 0.954 & \underline{32.80} & 0.931 & 30.99 & 0.906 & 30.92 & 0.908 \\
ChaIR~\cite{cui2023chair}              & KBS2023   & 28.19 & 0.879 & 28.69 & 0.862 & 34.52 & 0.953 & \textbf{32.85} & 0.931 & 31.30 & 0.903 & 31.11 & 0.906 \\
IRNeXT~\cite{cui2023irnext}            & ICML2023  & 25.80 & 0.860 & 27.22 & 0.833 & 31.65 & 0.931 & 30.53 & 0.917 & 29.02 & 0.898 & 28.85 & 0.888 \\
OKNet~\cite{cui2024oknet}              & AAAI2024  & 25.43 & 0.858 & 24.01 & 0.804 & 31.19 & 0.928 & 29.32 & 0.911 & 27.56 & 0.886 & 27.50 & 0.877 \\
AST~\cite{zhou2024ast}                 & CVPR2024  & 26.07 & 0.859 & 27.40 & 0.833 & 32.03 & 0.932 & 31.65 & 0.921 & 30.69 & 0.897 & 29.57 & 0.889 \\
SFHformer~\cite{jiang2024sfhformer}   & ECCV2024  & 25.67 & 0.856 & 27.25 & 0.832 & 32.97 & 0.944 & 32.27 & 0.925 & 31.50 & 0.904 & 29.94 & 0.892 \\
Nerd-rain~\cite{chen2024bidirectional} & CVPR2024  & 27.16 & 0.869 & 28.07 & 0.838 & 33.72 & 0.949 & 32.63 & 0.927 & 30.45 & 0.890 & 30.41 & 0.895 \\
FSNet~\cite{cui2024fsnet}               & TPAMI2024 & 27.95 & 0.884 & 28.70 & 0.860 & 34.10 & 0.952 & 32.68 & 0.931 & 31.26 & 0.910 & 30.94 & 0.908 \\
DMSR~\cite{zou2025dmsr}                 & ICME2025  & 28.88 & 0.890 & 29.41 & 0.873 & 35.19 & 0.957 & 32.50 & 0.931 & 31.35 & 0.910 & 31.47 & 0.912 \\
AdaIR~\cite{cui2025adair}               & ICLR2025  & 28.64 & 0.889 & 29.48 & 0.871 & \textbf{35.84} & \textbf{0.962} & 32.70 & 0.930 & 30.58 & 0.907 & 31.45 & 0.912 \\
\midrule
AFRNet-S (Ours) & --
  & \underline{29.20} & \underline{0.900}
  & \underline{29.75} & \underline{0.885}
   & 35.36 & 0.960
  & 32.73 & \underline{0.937}
  & \underline{31.60} & \underline{0.917}
  & \underline{31.73} & \underline{0.920} \\
AFRNet-L (Ours) & --
  & \textbf{29.41} & \textbf{0.906}
  & \textbf{29.98} & \textbf{0.893}
  & \underline{35.43} & \underline{0.961}
  & \textbf{32.85} & \textbf{0.940}
  & \textbf{31.73} & \textbf{0.921}
  & \textbf{31.88} & \textbf{0.924} \\
\bottomrule
\end{tabular*}
\end{table*}

\begin{table*}[!t]
\caption{Quantitative Comparison on Rain200L, Rain200H, and SPA-Data in
  Terms of Y-Channel PSNR and SSIM. ``--'' denotes an unavailable result.
  Best and second-best results are shown in \textbf{bold} and
  \underline{underlined}, respectively.}
\label{tab:rain200_spa}
\centering
\footnotesize
\setlength{\tabcolsep}{1.0pt}
\begin{tabular*}{\tblwidth}{@{\extracolsep{\fill}}L*{16}{C}@{}}
\toprule
Dataset
   & \multicolumn{2}{c}{MPRNet~\cite{zamir2021multi}}
   & \multicolumn{2}{c}{Uformer~\cite{wang2022uformer}}
   & \multicolumn{2}{c}{Restormer~\cite{zamir2022restormer}}
   & \multicolumn{2}{c}{DRSformer~\cite{Chen_2023_CVPR_DRSformer}}
   & \multicolumn{2}{c}{MambaIR~\cite{guo2024mambair}}
   & \multicolumn{2}{c}{DFSSM~\cite{yamashita2024dfssm}}
  & \multicolumn{2}{c}{AFRNet-S}
  & \multicolumn{2}{c}{AFRNet-L} \\
\cmidrule(lr){2-3}
\cmidrule(lr){4-5}
\cmidrule(lr){6-7}
\cmidrule(lr){8-9}
\cmidrule(lr){10-11}
\cmidrule(lr){12-13}
\cmidrule(lr){14-15}
\cmidrule(lr){16-17}
  & PSNR $\uparrow$ & SSIM $\uparrow$
  & PSNR $\uparrow$ & SSIM $\uparrow$
  & PSNR $\uparrow$ & SSIM $\uparrow$
  & PSNR $\uparrow$ & SSIM $\uparrow$
  & PSNR $\uparrow$ & SSIM $\uparrow$
  & PSNR $\uparrow$ & SSIM $\uparrow$
  & PSNR $\uparrow$ & SSIM $\uparrow$
  & PSNR $\uparrow$ & SSIM $\uparrow$ \\
\midrule
Rain200L
   & 39.47 & 0.983
   & 40.20 & 0.986
   & 40.99 & 0.989
   & 41.23 & 0.989
   & 41.13 & 0.990
   & 41.81 & \textbf{0.991}
   & \underline{42.28} & 0.990
   & \textbf{42.52} & \textbf{0.991} \\
Rain200H
   & 30.67 & 0.911
   & 30.80 & 0.911
   & 32.00 & 0.933
   & 32.17 & 0.933
   & 32.18 & 0.930
   & \underline{32.99} & \underline{0.940}
    & 32.90 & \underline{0.940}
  & \textbf{33.02} & \textbf{0.944} \\
\midrule
SPA-Data
   & 43.64 & 0.984
   & 46.13 & 0.991
   & 47.98 & 0.992
   & 48.54 & 0.992
  & -- & --
   & 49.55 & \underline{0.994}
  & \underline{49.83} & \underline{0.994}
  & \textbf{49.90} & \textbf{0.995} \\
\bottomrule
\end{tabular*}
\end{table*}

\begin{table}[!t]
\caption{Quantitative Comparison on the Official RainDS-Syn 3,000/600 Split.
  Each entry presents Y-channel PSNR/SSIM; the three columns correspond to
  streak (RS), raindrop (RD), and compound-rain (RDRS) degradation.}
\label{tab:rainds_syn}
\centering
\scriptsize
\setlength{\tabcolsep}{1.4pt}
\begin{tabular*}{\tblwidth}{@{\extracolsep{\fill}}LLCCC@{}}
\toprule
Method & Venue & RS & RD & RDRS \\
\midrule
MPRNet~\cite{zamir2021multi}       & CVPR2021     & 40.81/0.981 & 37.03/0.972 & 34.79/0.956 \\
DGUNet~\cite{mou2022dgunet}        & CVPR2022     & 41.09/0.983 & 37.56/0.975 & 35.34/0.959 \\
Uformer~\cite{wang2022uformer}     & CVPR2022     & 40.69/0.972 & 37.08/0.966 & 34.99/0.954 \\
Restormer~\cite{zamir2022restormer} & CVPR2022 (Oral) & 41.42/0.980 & 38.78/0.976 & 36.08/0.961 \\
IDT~\cite{xiao2022idt}             & CVPR2022     & 41.61/0.983 & 39.09/0.980 & 36.23/0.960 \\
NAFNet~\cite{chen2022nafnet}       & ECCV2022     & 40.39/0.972 & 37.23/0.974 & 34.99/0.957 \\
UDR-S2Former~\cite{chen2023udrs2former} & ICCV2023 & \underline{42.39}/\underline{0.988} & \underline{39.78}/\textbf{0.983} & \underline{36.91}/\underline{0.966} \\
\midrule
AFRNet (Ours)                    & Ours & \textbf{42.70}/\textbf{0.991} & \textbf{39.81}/\textbf{0.983} & \textbf{37.12}/\textbf{0.970} \\
\bottomrule
\end{tabular*}
\end{table}

Under the Rain13K protocol, Table~\ref{tab:rain13k} shows that AFRNet-L
obtains the best average performance and ranks first or tied for first on four
of the five test sets, with AdaIR remaining stronger on Rain100L. AFRNet-S
achieves the second-best average score and remains competitive on four test
sets; on Test2800, it ranks fourth in PSNR. The compact configuration
nevertheless preserves most of the performance of the larger model. The
improvements are also reflected in the restored appearance: compared with
the competing outputs, fewer streak remnants remain in the enlarged regions,
while object contours and local textures are less affected by over-smoothing.
This behavior is observed on both light- and heavy-rain test sets, suggesting
that the gain is not restricted to a particular rain density.

Table~\ref{tab:rain200_spa} shows that the model is effective across different
rain densities and real rainy scenes. AFRNet-L achieves the best PSNR and SSIM
on all three benchmarks, while AFRNet-S obtains the second-highest PSNR on
Rain200L and the third-highest PSNR on Rain200H. On Rain200L/H, the corresponding
reconstructions remove dense streak structures while retaining thin
boundaries and background textures. On SPA-Data, the visual comparisons show
cleaner scene regions with sharper architectural edges and local textures,
which is consistent with its higher SSIM.

RainDS evaluates different degradation types within a common protocol.
Table~\ref{tab:rainds_syn} shows that AFRNet obtains the highest PSNR on RS,
RD, and RDRS, and achieves the best or tied for the best SSIM on all three
subsets. For
streak degradation, the restored images contain fewer line-like artifacts
without visibly removing the main scene structures. For raindrop
degradation, broad drop-shaped distortions are suppressed while windows,
facade contours, and other background details remain recognizable. Under
compound rain, both streak and drop artifacts are reduced simultaneously,
which explains the consistent performance across the three degradation
types.

\subsection{Qualitative Results}

\begin{figure*}[!t]
  \centering
  \safeincludegraphics[width=\textwidth]{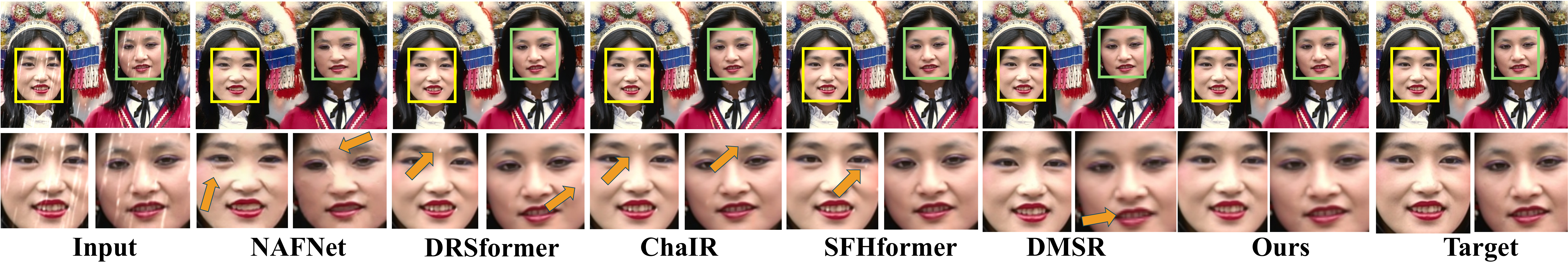}
\caption{Qualitative comparison on Rain100L. Enlarged regions highlight
  residual streak suppression and fine-detail preservation.}
  \label{fig:comparison_l}
\end{figure*}

\begin{figure*}[!t]
  \centering
  \safeincludegraphics[width=\textwidth]{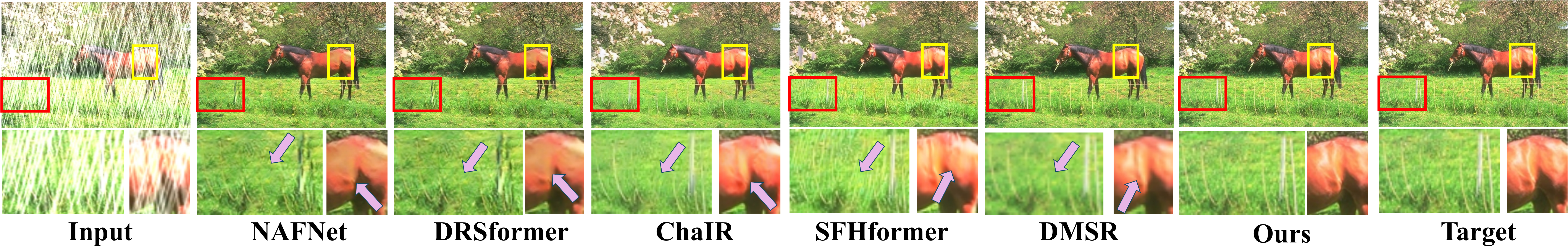}
\caption{Qualitative comparison on Rain100H under heavy-rain conditions.
  Enlarged regions highlight streak removal and structure preservation.}
  \label{fig:comparison_h}
\end{figure*}

Figs.~\ref{fig:comparison_l} and~\ref{fig:comparison_h} compare AFRNet with
representative methods on Rain100L and Rain100H. Under light rain, AFRNet
removes the remaining streaks in the enlarged regions while preserving thin
structures. Under heavy rain, it produces cleaner building edges and fewer
streak remnants than the competing methods, which either retain visible rain
responses or smooth the underlying details.

\begin{figure*}[!t]
  \centering
  \safeincludegraphics[width=\textwidth]{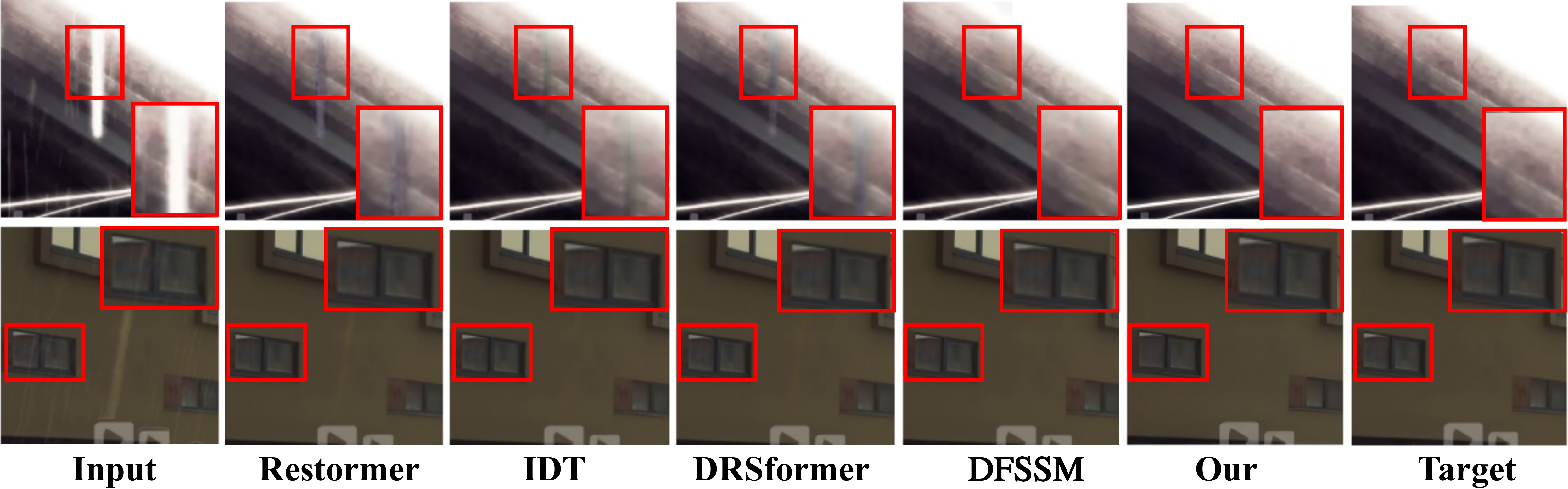}
\caption{Qualitative comparison on SPA-Data. Enlarged regions highlight
  streak suppression and edge preservation in real rainy scenes.}
  \label{fig:spa_data_vis}
\end{figure*}

Fig.~\ref{fig:spa_data_vis} shows the comparison on SPA-Data. AFRNet reduces
the visible rain haze while preserving local contrast around building
boundaries and vegetation. Compared with the competing outputs, the restored
regions contain sharper edges and fewer residual rain responses. Table~\ref{tab:rain200_spa}
reports the corresponding improvements in PSNR and SSIM.

\begin{figure*}[!t]
  \centering
  \safeincludegraphics[width=.958\textwidth]{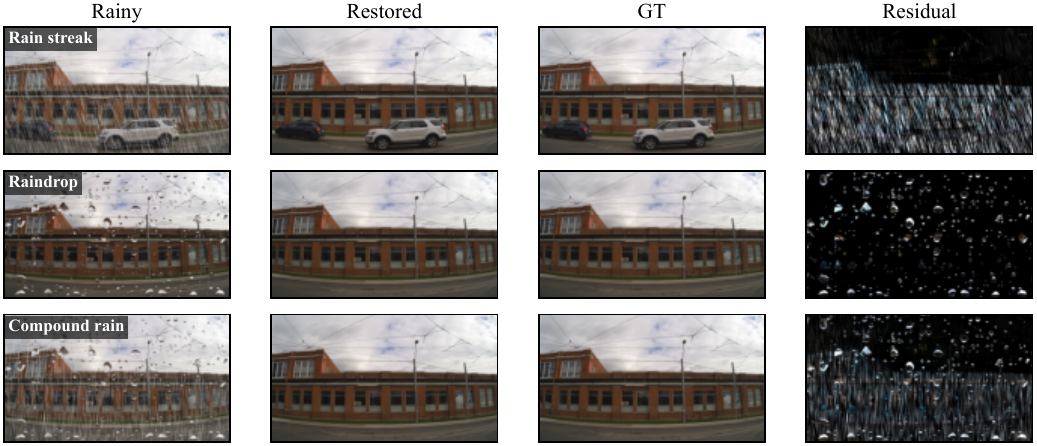}
\caption{Qualitative results on RainDS-Syn under streak, raindrop, and
  compound-rain degradation types. Residual maps visualize the removed rain
  components.}
  \label{fig:rainds_syn_vis}
\end{figure*}

\begin{figure*}[!t]
  \centering
  \safeincludegraphics[width=.958\textwidth]{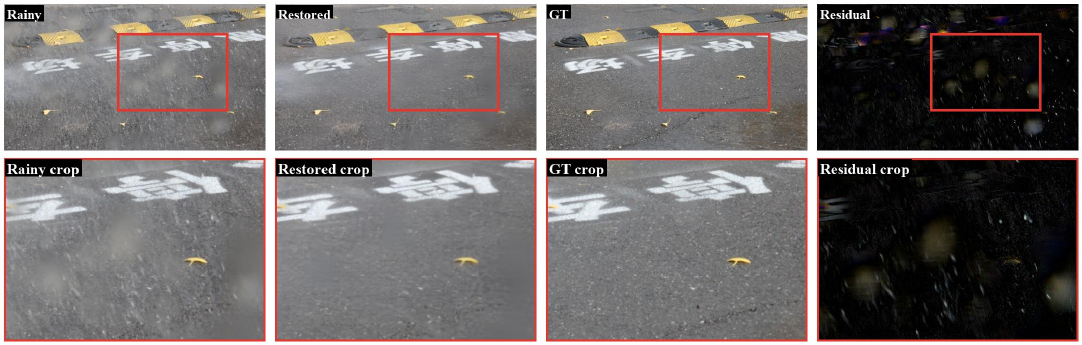}
\caption{Qualitative results on unpaired RainDS-Real images.}
  \label{fig:rainds_real_vis}
\end{figure*}

Figs.~\ref{fig:rainds_syn_vis} and~\ref{fig:rainds_real_vis} extend the
comparison to synthetic and real rainy images. On RainDS-Syn, the residual
maps are concentrated on the visible streak and drop regions, while the
restored images retain the scene layout and surface details. On the unpaired
RainDS-Real examples, AFRNet reduces haze and drop-like artifacts without
removing road markings or other scene structures. These results provide
qualitative evidence of robustness on real rainy images.

\subsection{Transferability to Image Deblurring}
\label{sec:gopro}

To assess transferability beyond deraining, we further train AFRNet on the
GoPro image deblurring benchmark~\cite{nah2017gopro}. We use $256\times256$
training crops and the same 300K-iteration schedule, augmentation, optimizer,
and loss as above, and report RGB PSNR and SSIM. AFRNet is evaluated under
our training protocol.

\begin{table}[!t]
\caption{Quantitative comparison on GoPro image deblurring in terms of RGB
  PSNR and SSIM. The best results are shown in bold.}
\label{tab:gopro}
\centering
\scriptsize
\setlength{\tabcolsep}{3.0pt}
\begin{tabular*}{\tblwidth}{@{\extracolsep{\fill}}LCC@{}}
\toprule
Method & PSNR $\uparrow$ & SSIM $\uparrow$ \\
\midrule
DMPHN~\cite{zhang2019dmphn} & 31.20 & 0.940 \\
Restormer~\cite{zamir2022restormer} & 32.92 & 0.961 \\
SFNet~\cite{cui2023sfnet} & 33.27 & 0.963 \\
ACL~\cite{gu2025acl} & 33.25 & 0.964 \\
AFRNet (Ours) & \textbf{33.57} & \textbf{0.968} \\
\bottomrule
\end{tabular*}
\end{table}

Table~\ref{tab:gopro} shows that AFRNet achieves the highest PSNR and SSIM
among the listed methods, with 33.57 dB and 0.968, respectively. This result
indicates that the proposed frequency decomposition and subband processing
remain effective for motion blur. It also demonstrates the transferability of the proposed restoration design
to image deblurring, without making a claim of state-of-the-art performance on
this task.

\begin{figure}[!t]
  \centering
  \safeincludegraphics[width=\columnwidth]{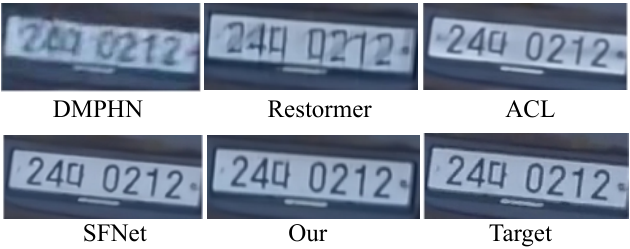}
  \caption{Qualitative comparison on GoPro image deblurring.}
  \label{fig:gopro}
\end{figure}

Fig.~\ref{fig:gopro} compares AFRNet with DMPHN~\cite{zhang2019dmphn},
SFNet~\cite{cui2023sfnet}, ACL~\cite{gu2025acl}, and Restormer~\cite{zamir2022restormer}.
AFRNet produces clearer local edges and text-like details with fewer residual
blur artifacts in the selected examples, which is consistent with its
33.57/0.968 result.

\subsection{Efficiency Analysis}

\begin{table}[!t]
\caption{Model Complexity for a $256\!\times\!256$ Input. ``--'' denotes an
  unavailable profile.}
\label{tab:complexity}
\begin{tabular*}{\tblwidth}{@{\extracolsep{\fill}}LCC@{}}
\toprule
Method & Params (M) $\downarrow$ & GFLOPs $\downarrow$ \\
\midrule
DRSformer~\cite{Chen_2023_CVPR_DRSformer} & 33.7 & 242.9 \\
TransWeather~\cite{valanarasu2022transweather} & 31.0 & -- \\
AdaIR~\cite{cui2025adair} & 28.77 & 147.45 \\
UDR-S2Former~\cite{chen2023udrs2former} & 28.53 & 21.58 \\
Restormer~\cite{zamir2022restormer} & 26.12 & 174.7 \\
NeRD-Rain~\cite{chen2024bidirectional} & 22.89 & 156.3 \\
FourierMamba~\cite{li2024fouriermamba} & 17.62 & 22.56 \\
FreqMamba~\cite{zou2024freqmamba} & 14.52 & 36.49 \\
MPRNet~\cite{zamir2021multi} & 3.64 & 148.55 \\
\midrule
AFRNet-S (Ours) & 2.46 & 64.90 \\
AFRNet-L (Ours) & 3.88 & 91.16 \\
\bottomrule
\end{tabular*}
\end{table}

Table~\ref{tab:complexity} shows that AFRNet-S uses 2.46\,M parameters and
64.90\,GFLOPs, while AFRNet-L uses 3.88\,M parameters and 91.16\,GFLOPs.
AFRNet-S is substantially smaller than DRSformer and Restormer while using
less computation than these models. AFRNet-L also remains compact relative to
the larger baselines and provides a higher-capacity configuration of the
same architecture.

\subsection{Ablation Study}
\label{sec:ablation}

We conduct the ablation study on Rain200H under the same training protocol.
Table~\ref{tab:ablation} isolates the contributions of AFRM and SCB and
compares their simplified and complete forms.

\begin{figure*}[!t]
  \centering
  \safeincludegraphics[width=\textwidth]{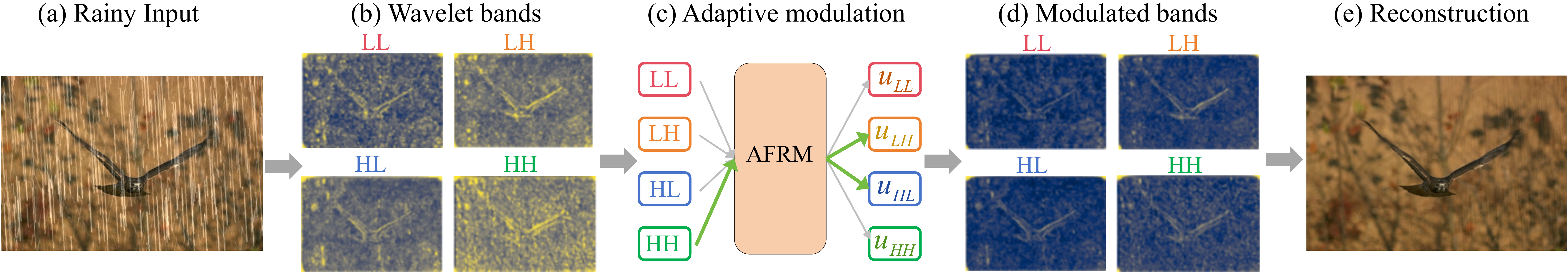}
  \caption{Visualization of the wavelet-domain operation of AFRM. (a) Input
    rainy image. (b) Wavelet subbands of the input image. (c) Content-adaptive
    modulation by AFRM. (d) Modulated wavelet subbands. (e) Image reconstructed
    from the modulated subbands.}
  \label{fig:ablation_afrm}
\end{figure*}

\begin{table}[!t]
\caption{Ablation Study on Rain200H.}
\label{tab:ablation}
\centering
\scriptsize
\setlength{\tabcolsep}{1.0pt}
\begin{tabular*}{\tblwidth}{@{\extracolsep{\fill}}LCCCC@{}}
\toprule
Model & AFRM & SCB & PSNR $\uparrow$ & SSIM $\uparrow$ \\
\midrule
 Backbone (baseline) & $\xmark$ & $\xmark$ & 31.25 & 0.928 \\
 + AFRM (within-band modulation) & within-band & $\xmark$ & 31.38 & 0.930 \\
 + AFRM (full cross-band mixing) & $\cmark$ & $\xmark$ & 32.01 & 0.934 \\
 + SCB (independent $\beta$) & $\xmark$ & independent $\beta$ & 31.49 & 0.931 \\
 + SCB (full compensation) & $\xmark$ & $\cmark$ & 32.14 & 0.935 \\
 AFRNet (Full) & $\cmark$ & $\cmark$ & 33.02 & 0.944 \\
\bottomrule
\end{tabular*}
\end{table}

Table~\ref{tab:ablation} shows that the simplified variants provide only
limited improvements over the backbone. Within-band AFRM reaches 31.38/0.930
and independent SCB reaches 31.49/0.931 in PSNR/SSIM, compared with
31.25/0.928 for the backbone.
With complete cross-subband mixing, AFRM reaches 32.01/0.934, corresponding
to a $0.76$\,dB PSNR gain over the backbone. This gap indicates that the
  benefit of AFRM comes from cross-subband interaction rather than from
  within-band modulation alone. Full SCB reaches 32.14/0.935, yielding a
$0.89$\,dB gain over the backbone and a larger standalone improvement than
 AFRM. When both complete modules are enabled, AFRNet achieves 33.02/0.944,
which is a further $1.01$\,dB improvement over the
backbone + AFRM configuration. The result supports the complementary roles of
cross-subband rain suppression and decoder-detail compensation.

Fig.~\ref{fig:ablation_afrm} shows that the input image is first decomposed
into the four wavelet subbands $LL$, $LH$, $HL$, and $HH$, which are then
modulated by AFRM. In this example, the energy in the $HH$ subband is
redistributed toward the $LH$ and $HL$ subbands. The resulting modulated
subbands produce a reconstructed image with substantially fewer residual rain
streaks, while the main scene content is retained. Table~\ref{tab:ablation}
reports the larger gain obtained from full cross-subband mixing.

\begin{figure}[!t]
  \centering
  \safeincludegraphics[width=\columnwidth]{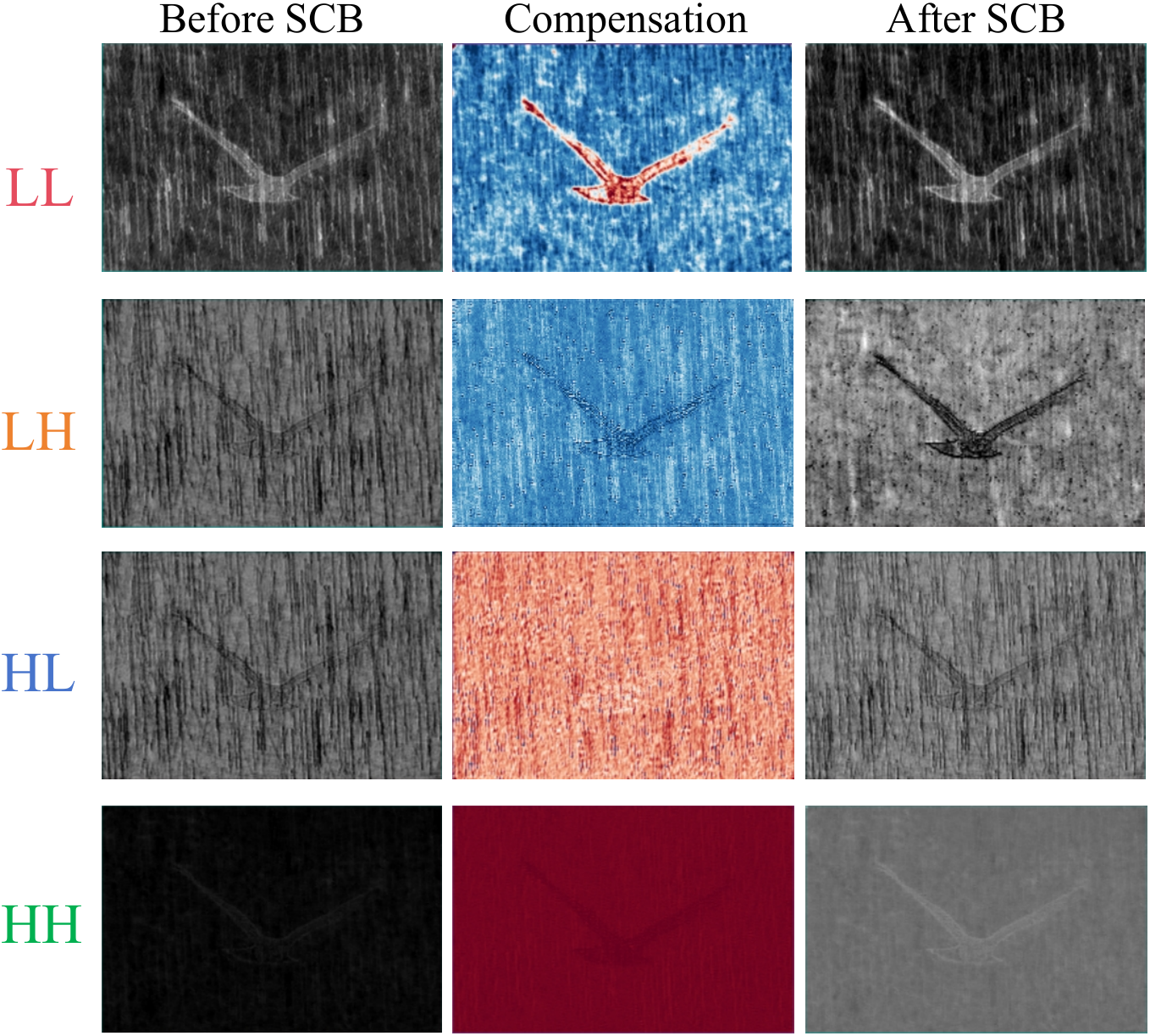}
  \caption{Visualization of SCB compensation across wavelet subbands. The
    four rows correspond to the $LL$, $LH$, $HL$, and $HH$ subbands,
    respectively. From left to right, the columns show the subband before
    SCB, the compensation produced by SCB, and the subband after SCB.}
  \label{fig:ablation_scb}
\end{figure}

Unlike AFRM, SCB primarily restores lost details through subband-dependent
compensation. Fig.~\ref{fig:ablation_scb} shows the compensation process
across the four wavelet subbands. The four rows correspond to the four wavelet
subbands. After SCB
compensation, the contours and fine structures in the subbands become clearer,
although the compensation patterns differ across frequency bands. Table~\ref{tab:ablation}
shows that full SCB provides a stronger gain over the backbone.

\section{Future Work}

Future work will investigate finer frequency partitions and content-adaptive
subband routing for complex rain patterns. We will also explore transferring
AFRNet to all-in-one image restoration, where one model handles multiple
degradation types. Improving the efficiency of Fourier operations on
resource-constrained devices is another practical direction.

\section{Conclusion}

We have presented the Adaptive Frequency Redistribution Network (AFRNet) for
single-image deraining. We observed that
rain artifacts and clean structures may spread across correlated wavelet
subbands, so AFRNet introduces an Adaptive Frequency Recombination Module
(AFRM) to adaptively modulate and recombine them. To recover subband details
lost during encoder-decoder resolution changes, AFRNet further uses a Subband
Compensation Block (SCB) based on inter-subband attention to transfer encoder
information to decoder subbands. Experiments on synthetic, real-rain, and
compound-rain benchmarks show that AFRNet suppresses rain artifacts while
preserving image structures and textures; an additional GoPro image deblurring
experiment provides evidence of its generalization beyond deraining.

\bibliographystyle{IEEEtran}
\bibliography{refs}

\end{document}